\documentclass[10pt,twocolumn,letterpaper]{article}

\usepackage{iccv}
\usepackage{times}
\usepackage{epsfig}
\usepackage{graphicx}
\usepackage{amsmath}
\usepackage{amssymb}

\usepackage{color}
\usepackage{multirow}
\usepackage{graphicx}
\usepackage{amsmath}
\usepackage{amssymb}
\usepackage{booktabs}
\usepackage{array}
\usepackage{tikz}
\usepackage{multirow}
\usepackage[pagebackref,breaklinks,colorlinks]{hyperref}
\usepackage[capitalize]{cleveref}

\Crefname{section}{Section}{Sections}
\Crefname{table}{Table}{Tables}
\iccvfinalcopy 

\ificcvfinal\fi

\begin{document}

\title{Feature-domain Adaptive Contrastive Distillation for Efficient Single Image Super-Resolution}

\author{HyeonCheol Moon, Jinwoo Jeong, and Sungjei Kim \\
Korea Electronics Technology Institutes\\
Seongnam-si, Republic of Korea\\
{\tt\small \{hcmoon23, jinwoo.jeong, and sungjei.kim\}@keti.re.kr}
}

\ificcvfinal\thispagestyle{empty}\fi

\maketitle

\begin{abstract}
Recently, CNN-based SISR has numerous parameters and high computational cost to achieve better performance, limiting its applicability to resource-constrained devices such as mobile. As one of the methods to make the network efficient, Knowledge Distillation (KD), which transfers teacher's useful knowledge to student, is currently being studied. More recently, KD for SISR utilizes Feature Distillation (FD) to minimize the Euclidean distance loss of feature maps between teacher and student networks, but it does not sufficiently consider how to effectively and meaningfully deliver knowledge from teacher to improve the student performance at given network capacity constraints. In this paper, we propose a feature-domain adaptive contrastive distillation (FACD) method for efficiently training lightweight student SISR networks. We show the limitations of the existing FD methods using Euclidean distance loss, and propose a feature-domain contrastive loss that makes a student network learn richer information from the teacher's representation in the feature domain. In addition, we propose an adaptive distillation that selectively applies distillation depending on the conditions of the training patches. The experimental results show that the student EDSR and RCAN networks with the proposed FACD scheme improves not only the PSNR of the entire benchmark datasets and scales, but also the subjective image quality compared to the conventional FD approaches.

\end{abstract}

\section{Introduction}
\label{sec:intro}

Single image super-resolution (SISR) is a method of generating a high-resolution image from a given low-resolution image \cite{sr}. It is an important task that can be applied to a variety of computer vision tasks, such as medical imaging \cite{medical_image_2}, pattern and object recognition \cite{detection_kd, multimedia_1}. In the prior works, interpolation and example-based methods have been applied to the SISR task \cite{rep_based_2, rep_based, interpolation}. However, both methods show performance limitations. Relatively recently, the advent of CNN-based SISR networks such as SRCNN \cite{srcnn} has provided outperformed the traditional SISR works. Since then, numerous CNN-based SISR networks have been proposed \cite{sr, vdsr, drcn}, and the network parameters and computational complexity have been increased to obtain better performance. 

\begin{figure}
\centering
\includegraphics[width=8cm]{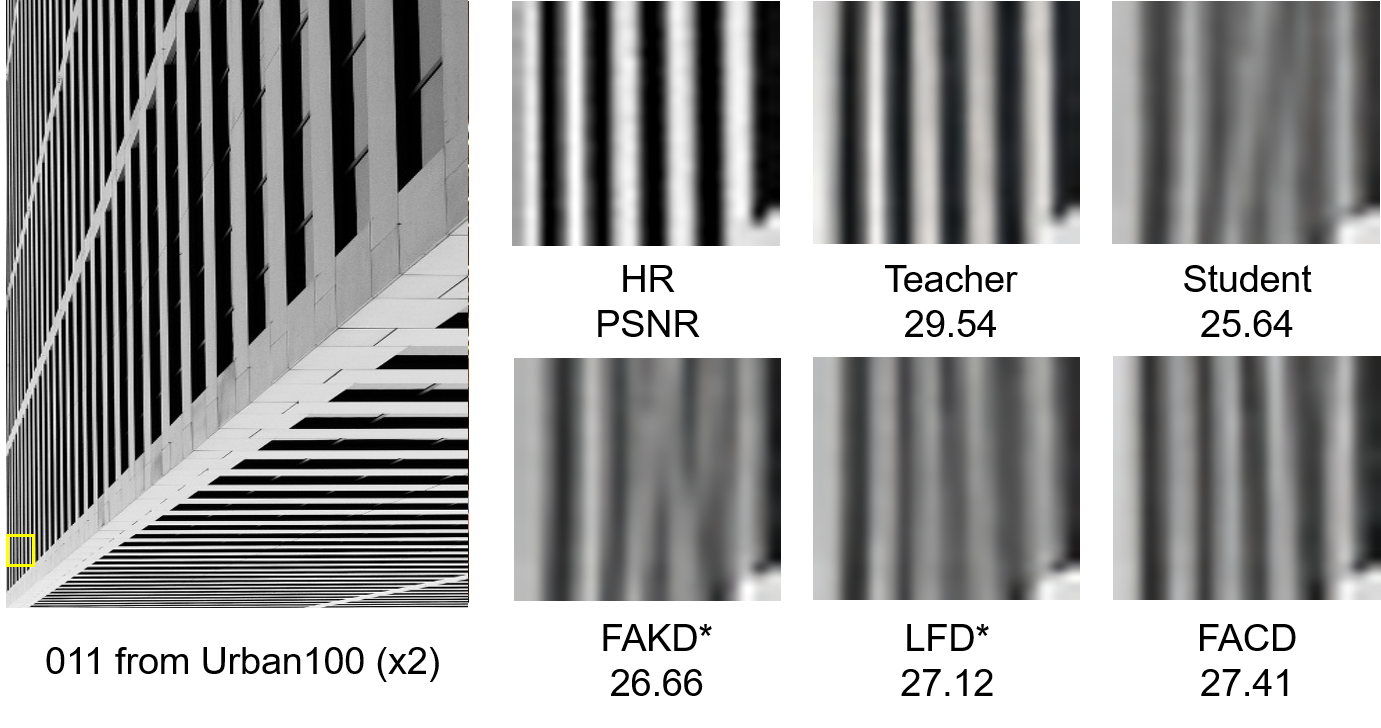}
\caption{Examples of limitations on FD with Euclidean loss \cite{fakd, lsfd}. The teacher and student networks are student EDSR \cite{edsr2} with scale x2 image. Noted that FAKD* and LFD* indicate our reproduced results with the same experimental settings. \label{figure_0}}
\end{figure}

The complex SISR model has limitations in practical applications such as resource-constrained devices. For the resource-limited environment such as mobile or IoT devices, the efficient and lightweight SISR model is needed. To meet this demand, the lightweight SISR model which derives more efficient performance trade-off is being studied \cite{fakd,f_pruning2, ncnet, lsfd, csd2021, f_pruning,assl, srp}. Among the above-mentioned methods, KD \cite{kd} methods have the following distinctive advantages: 1) KD has the advantage of inheriting the knowledge of large teachers and improving performance without modifying the network structure which is already in place at the industrial site, 2) KD can be combined with pruning and network design methods by adding loss terms to achieve more performance improvement \cite{combine, pqk}.

KD is mainly used for classification and detection tasks \cite{detection_kd, crd, detect_kd}. The student network is trained to minimize the distance between the labels of the student network and the soft labels of the teacher network in the classification task. However, this approach for SISR task shows a limitation in performance improvement \cite{lsfd, fitnet}. In order to solve the problem, Feature domain Distillation (FD) is additionally introduced to guide the training of the student network. Feature Affinity-based KD (FAKD) \cite{fakd} proposed to transfer the intermediate feature knowledge of a larger teacher model to a lightweight student network. FAKD found that FD with image domain distillation helped to improve the distillation performance. After that, Local Feature Distillation/Local-Selective Feature Distillation (LFD/LSFD) \cite{lsfd} proposed the feature attention method, which selectively focuses on specific positions to extract refined feature information, to improve the simple distance-based feature distillation of FAKD. Both methods have in common that they use Euclidean distance as a metric to transfer the feature knowledge from teacher to student. As shown in \cref{figure_0}, neither method completely solves the disadvantages of Euclidean distance-based loss such as the pattern loss and image blurring. 

\begin{figure}
\centering
\includegraphics[width=8cm]{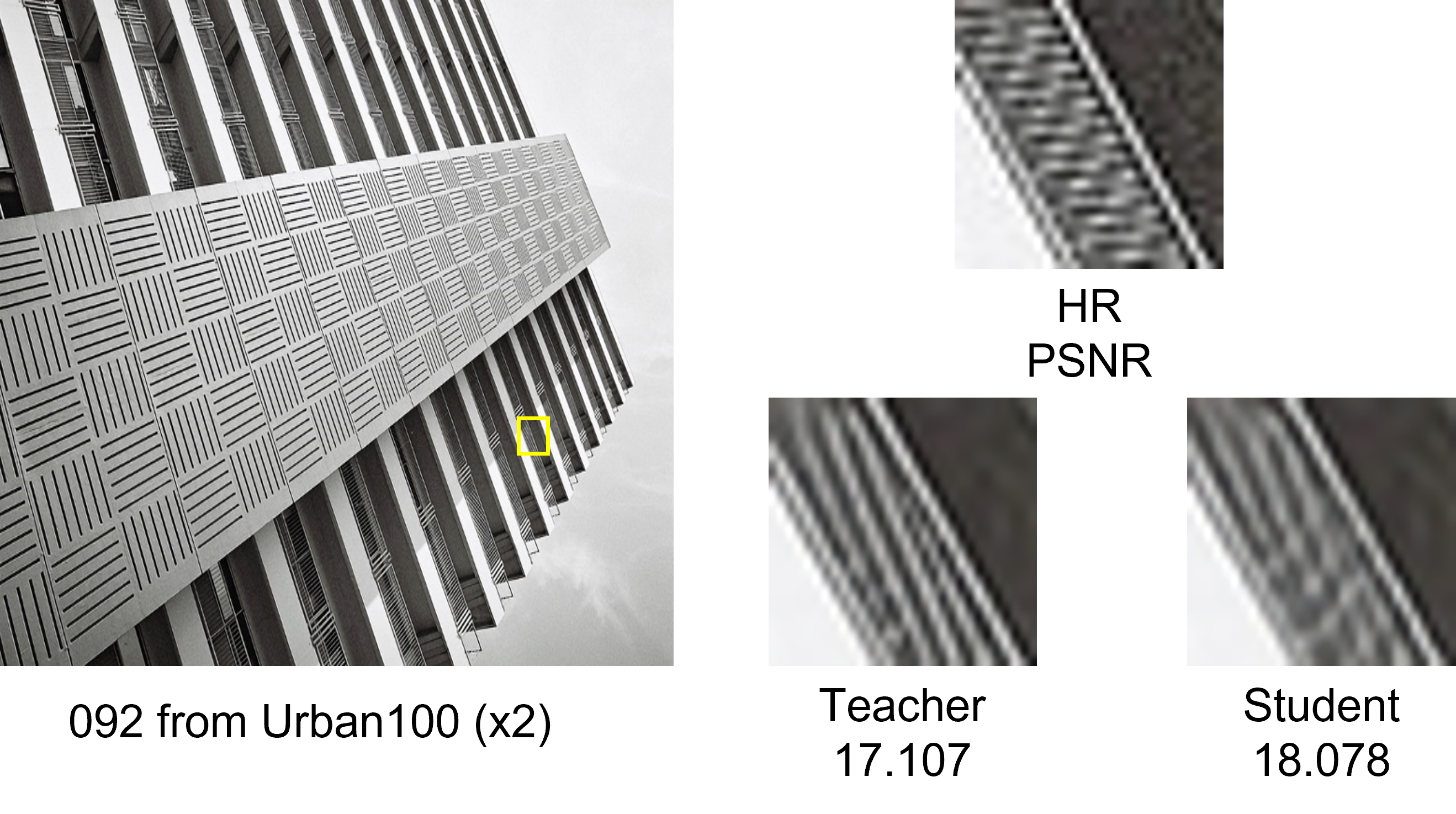}
\caption{Examples of worse teacher case on EDSR with x2 SR. Noted that teacher and student networks are trained separately from scratch.\label{figure_1}}
\end{figure}

To overcome the limitations of Euclidean distance loss, contrastive loss for SISR has been studied in the KD scheme \cite{srcrd, csd2021}. Contrastive Self-Distillation (CSD) \cite{csd2021} proposed the scheme to explicitly transfer the knowledge from teacher to student using contrastive loss in the latent space of image domain and improved the distillation performance and texture restoration. However, in the SISR, this approach can reduce the efficiency of distillation because it cannot fully exploit the rich information of intermediate feature maps. Moreover, as shown in \cref{figure_1}, the inference output of teacher network does not guarantee better performance for all patches. Those inappropriate inference results interfere with the training of the student network and need to be removed from the training process.

To address these issues, we propose a Feature-domain Adaptive Contrastive Distillation (FACD), which selectively transfers the teacher's feature-domain knowledge with contrastive loss. Our Feature-domain Contrastive Distillation (FCD) can solve the restoration of edges and patterns, and provide an improvement of the distillation performance compared to the Image-domain Contrastive Distillation (ICD) and CSD \cite{csd2021} by transferring well-refined feature knowledge to student effectively.

In addition, FAKD and LSFD, which use three intermediate feature maps for FD, do not account for attention at the feature map level. Since CNN-based SR networks have a cascading structure, we found that improper distillation at the top sequentially affects the bottom of the network. To this end, we assign the higher importance (attention) to the top of network and demonstrate it in the \cref{abl}.

Finally, the percentage of inappropriate inference output of the teacher network is up to 5\% for Enhanced Deep Residual Networks (EDSR) \cite{edsr2} and up to 11\% for Residual Channel Attention Networks (RCAN) \cite{RCAN} network across all training patches. Since these patches can transfer incorrect knowledge to the student network, we adaptively adjust whether to apply FCD based on patch conditions during training. Combined with feature-domain contrastive loss, feature map level attention, and adaptive distillation, FACD achieves state-of-the-art performance over FD and excellent qualitative results.
Our main contributions can be summarized as follows:
\begin{enumerate}
\item We propose an algorithm called FCD that improves the efficiency of traditional FD and mitigates the loss of detailed texture that occurs in the FD method through an intermediate feature domain contrastive learning method which can refine and transfer useful representational knowledge to students.
\item We found that inappropriate teacher's knowledge interfered with student learning. For the efficiency of distillation, we propose an algorithm called Feature-domain Adaptive Contrastive Distillation (FACD) that selectively applies FCD by comparing output patches derived from teacher and student networks with ground-truth. 
\item We demonstrate the FACD achieves state-of-the-art performance over distance-based FD and excellent qualitative results. In particular, in terms of qualitative results, FACD shows outperformed edge and pattern reconstruction results. Furthermore, we show a various experimental analysis with ablation studies.
\end{enumerate}

\section{Related Works}

\subsection{Efficient Super Resolution Network}
At first, the CNN-based SR model stacks deeper layers to improve performance, but this architecture caused a gradient vanishing. After that, VDSR \cite{vdsr} and DRCN \cite{drcn} networks used deep stacking of residual blocks \cite{rb} to solve this issue. In addition, EDSR \cite{edsr2} shows that Batch Normalization (BN) on the SISR model leads to normalization of the features, eliminating the flexibility of the model. Furthermore, EDSR uses the residual-scaling method to improve the instability of training due to the removal of BN. Recently, RCAN \cite{RCAN} and SAN \cite{san} provide significant performance improvement by adopting the channel-attention mechanism. 

However, deep layers, stacking blocks, and attention mechanisms require huge memory and computational costs for inference due to the large number of parameters, spatial and non-local operations, and are limited in their applicability to resource-constrained devices such as mobile or IoT devices. To adapt to these devices, it is essential to design an efficient network structure and optimize training schemes \cite{han}. Since optimizing performance by only designing an efficient network structure is limited, the advanced training schemes consisting of pruning, quantization, and KD become even more important on resource-constrained devices. Among them, KD is attractive because it can provide additional performance improvements without changing the structure of the target model, which is already applied in industry. The details of this approach are described in the next section.

\subsection{Feature domain Distillation for SISR}
KD is a method to transfer knowledge from the teacher model to a lightweight student networks \cite{kd}. Distillation with the label domain (same as the image domain in SISR) shows better performance in the classification problem. However, in the regression problems like SISR, the solution space is huge, so single image domain KD is not an effective way to transfer knowledge \cite{regress}. Therefore, FD mainly proposed to effectively guide the training of the student network through Euclidean distance-based matching in the image and feature domain \cite{fakd, lee, lsfd, fitnet}. 

First, FitNet \cite{fitnet} proposed the distillation in both image and feature domains. For FD, they use a simple regressor which is composed of \begin{math}1\times1\end{math} convolution layers due to the channel size of the teacher and student network is different. PISR \cite{lee} proposed to use ground-truth images as privileged information to make an encoder in the teacher network learn the degradation and subsampling of high-resolution images. For more efficient distillation, FAKD proposes a feature affinity-matrix-based KD framework by distilling the structural knowledge from a larger teacher model. Furthermore, the teacher supervision (TS) loss between the output super-resolution images of teacher and student is considered. In addition, LSFD proposed a feature attention method that adaptively focuses on the specific pixel to extract feature information by utilizing the difference map between the inference output of teacher and student. By merging FD with the adaptive functional attention mechanism, LSFD showed the performance enhancement compared to other FD algorithms such as FAKD. However, these approaches did not completely solve the limitation of the Euclidean distance loss in subjective image quality.

\subsection{Contrastive Learning}
Contrastive loss is mainly proposed in self-supervised learning \cite{contrastive, adv} and is used to train images so that positive pairs stay close to each other, while negative pairs are far away \cite{crd, csd2021}. By maximizing the Kullback-Leibler (KL) divergence of the positive and negative pairs, the mutual information in the positive pair is maximized while both distributions are clearly distinguished. In other words, KD with contrastive loss optimizes the performance by maximizing mutual information between teacher and student while minimizing uncertainty between both networks. This means that as training progresses, the networks of teacher and student become gradually similar. In particular, in this approach, contrastive representation distillation (CRD) achieves state-of-the-art (SOTA) results in the classification task by distillation with contrastive loss \cite{crd}. 

KD with contrastive loss \cite{srcrd, csd2021} in the image domain is proposed for SISR, and they have a slight performance improvement over other distillation approaches \cite{fakd, lsfd}. In particular, CSD \cite{csd2021} proposed to use contrastive loss in the latent features of the image domain. Inspired by conventional FD and CSD, we focus on the improving distillation by using contrastive loss in the feature domain where the solution space in the regression task is smaller than image domain. Therefore, we propose a novel method of feature-domain contrastive distillation and introduce an adaptive KD approach to efficiently transfer knowledge from teacher networks. 

\begin{figure*}
\centering
\includegraphics[width=16cm]{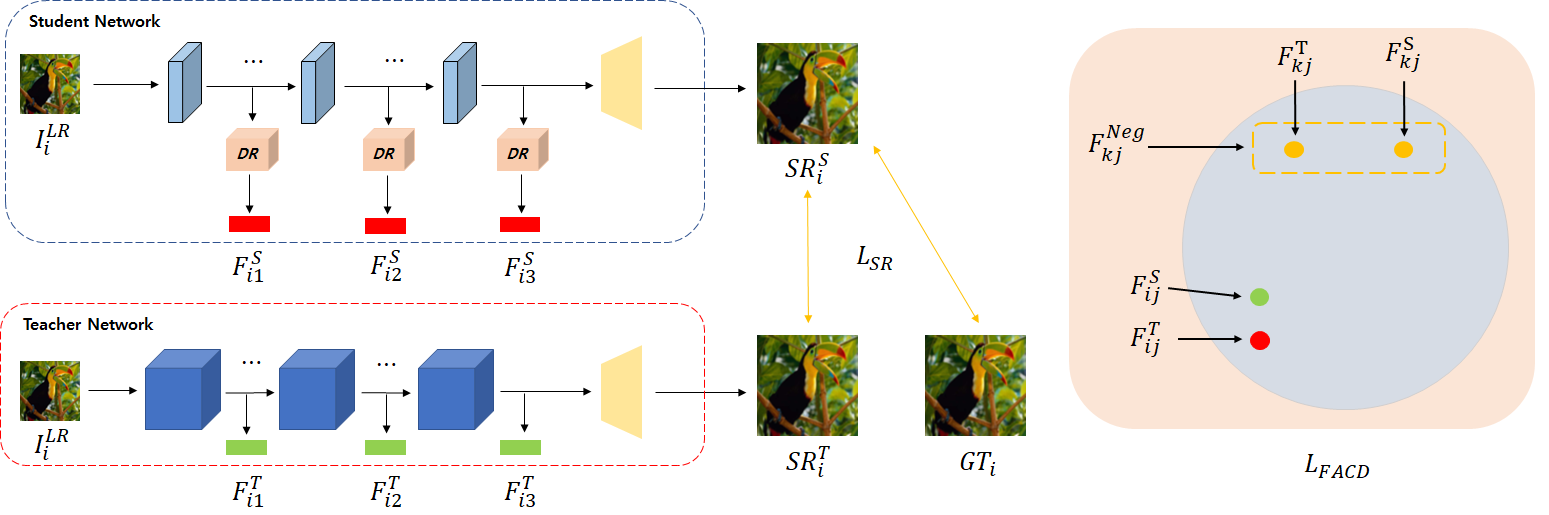}
\caption{Overall architecture of Feature-domain Adaptive Contrastive Distillation (FACD)\label{fig:framework}}
\end{figure*}
\section{Proposed Method}

In this section, we describe the loss function of the proposed distillation method. The pipeline of our proposed FACD framework is shown in \cref{fig:framework}. As shown in \cref{fig:framework}, the proposed FACD performs distillation on both image and feature domains. In the image domain, output images of the teacher and ground-truth (GT) are used to KD for the student network, respectively. On the other hand, in the feature domain, FACD performs with contrastive learning between the intermediate feature maps of the teacher and student \cite{fakd,lsfd}.

\subsection{Adaptive Knowledge Distillation for SR}
\label{adaptive}
In this section, we describe the adaptive KD for efficient knowledge transfer from teacher networks. As shown in \cref{figure_1}, interestingly, the output of the teacher network does not always guarantee better performance than students. We found that 5\% and 11\% on average of worse cases occur in EDSR and RCAN, respectively, in the training patches. These worse patches interfere with the efficiency of distillation. Therefore, we propose a simple but effective adaptive distillation method to optimize the distillation performance in both the image and feature domains. If the SR image of the student network is relatively closer to the ground truth than the SR image of the teacher network, we ignore these patches during training. The indicator of adaptive KD is formulated as:
\begin{equation}
  \alpha_i = \begin{cases} 0 & \text{if} \,\, \|SR_i^S-GT_i\|_1 < \|SR_i^T-GT_i\|_1, \\ 1 & \text{else}. \end{cases} \
  \label{eq:loss_aw}
\end{equation}
where \begin{math}\alpha\end{math} is the indicator of appropriate samples, and $i$ is the index of the batch sample. If the distance from GT is farther from the teacher, the parameter of appropriate samples \begin{math}\alpha_i\end{math} is set to 0. Here, 0 value means that the patch is not used for distillation.

\subsection{Contrastive Adaptive Distillation }
\label{fcd}
Conventional FD has demonstrated that the efficiency of distillation lies in the feature domain rather than the image domain \cite{fakd}. For this reason, previous FD methods for SISR have mainly focused on improving the distillation schemes in the feature domain. Nevertheless, the combination of feature distillation and image distillation has been shown to outperform a single refined feature distillation alone \cite{fakd, lsfd}. Moreover, KD with contrastive loss improves the performance by enabling more explicit knowledge transfer from teacher networks \cite{csd2021}. Therefore, we apply the contrastive learning in the feature-domain to transfer richer information from intermediate features, and keep the Euclidean distance loss in the image-domain distillation to minimize the interference from contrastive learning. We describe the loss function for each domain separately.

First, we propose the loss function in the image domain with the adaptive KD indicator (\begin{math}\alpha_i\end{math}) as follows:

\begin{equation}
\small
\begin{aligned}
  L_{SR}= \frac{1}{2N}{\sum_{i=1}^{N}{(2-\alpha_i)\|SR_i^S-GT_i\|_1} + {\alpha_i\|SR_i^S-SR_i^T\|_1}}
\end{aligned}
\label{eq:loss_sr}
\end{equation}

where \begin{math}SR^S\end{math}, \begin{math}SR^T\end{math}, and \begin{math}GT\end{math} are the output images of the student network, the teacher network, and ground-truth, respectively. \begin{math}N\end{math} is the number of batch size. The former of \begin{math}L_{SR}\end{math} means the Euclidean loss with GT, which is the loss term of the conventional SISR, and the latter means that distillation in the image domain.

Second, to transfer knowledge from intermediate features of teacher networks, we propose the feature domain adaptive contrastive distillation named FACD. For a fair comparison with the conventional FD methods \cite{fakd, lsfd}, FACD configured the three feature matching points as shown in \cref{fig:framework}. The detailed loss function of FACD is formulated as:

\begin{equation}
  \hat{F_{ij}} = \frac{F_{ij}}{\|F_{ij}\|_2}
  \label{eq:normal}
\end{equation}

\begin{equation}
  L_{FACD}=\sum_{i=1}^{N}{\sum_{j=1}^{3}{w_j\frac{\alpha_i\|DR(\hat{F_{ij}^S})-\hat{F_{ij}^T}\|_1}{\sum_{k=1}^{K}{\|DR(\hat{F_{ij}^S})-\hat{F_{kj}^{Neg}}\|_1}}}}
  \label{eq:loss_FACD}
\end{equation}

where \begin{math}\hat{F_{ij}}\end{math} is the normalized feature maps, \begin{math}w_j\end{math} refers to the attention weight of each feature matching point, and DR refers to the deep regressor consisting of five \begin{math}1\times1\end{math} convolutional layers with PReLU activation \cite{prelu}. In this paper, the feature map level attention parameter \begin{math}w_i\end{math} is set to [0.5, 0.3, 0.2]. In addition, N refers to the number of batch sizes, and K refers to the number of negative pairs. To better transfer knowledge from teacher, the feature maps are normalized. As with \cref{eq:loss_sr}, FD is not performed on inappropriate samples in positive pairs. 

In order to use the contrastive loss on feature domain, it is necessary to decide how to construct the positive and negative pairs, and which similarity measures (e.g. Euclidean, dot-product, and cosine similarity (CS)) to use in the contrastive loss function. 

For our FACD loss, as shown in \cref{fig:framework}, we consider the feature of the student network (\begin{math}F_{ij}^S\end{math}) and its feature of the teacher network (\begin{math}F_{ij}^T\end{math}) as a positive pair in the same index. To increase the efficiency of contrastive distillation, as shown in \cref{fig:framework}, all features of teacher and student except for the same index are considered as the negative pairs. To generate more negative samples, pairs with the different index on the student's features (\begin{math}F_{ij}^S\end{math}, \begin{math}F_{kj}^S\end{math}) are also considered as the negative pairs. Moreover, we adopted the contrastive loss with Euclidean loss as similarity measure. 

By minimizing this \begin{math}L_{FACD}\end{math}, the student network learns to place positive pairs closer and negative pairs further apart. Through this approach, the mutual information between the feature maps of the teacher and student networks can be maximized \cite{crd}. The effectiveness of contrastive loss on each domain is described in our ablation studies.

\subsection{Overall Loss Function}
\label{overall}
The overall loss function of our FACD is constructed by image and feature domain contrastive distillation, which can be formulated as \cref{eq:loss_total}:
\begin{equation}
  L_{total}= L_{SR}+ \lambda L_{FACD}
  \label{eq:loss_total}
\end{equation}
where \begin{math}\lambda\end{math} is a hyper-parameter for balancing \begin{math}L_{SR}\end{math} and   \begin{math}L_{FACD}\end{math}. The hyperparameter of the \begin{math}\lambda\end{math} is set to 4 in our experimental settings.

\section{Experiments}
In this section, we explain the details of our experimental network settings and analyze the experimental results both quantitatively and qualitatively.

\subsection{Experimental Settings}
Following the previous works \cite{san, fakd, edsr2, lsfd, csd2021, RCAN}, we use the 800 split set images from the DIV2K dataset \cite{div2k} for training. On the other hand, we test our FACD with Y-PSNR on four benchmark datasets such as Set 5 \cite{set5}, Set 14 \cite{set14}, BSD 100 \cite{b100}, and Urban 100 \cite{urban100}. For comparison with previous KD algorithms, we performed experiments on existing SISR networks, EDSR \cite{edsr2} and RCAN \cite{RCAN}. \cref{table:configuration} demonstrated the configuration of distillation models consisting of teacher and student networks. The configuration of the distillation model is the same as in the previous works for fair experimental comparison. While EDSR reduces the number of residual blocks (ResBlocks) and the channel size of the convolutional, RCAN keeps the number of residual groups (ResGroups) containing multiple ResBlocks, but only reduces the number of ResBlocks. This distillation compresses EDSR about 30 times and RCAN about 3 times in terms of the number of parameters.

Our FACD is implemented by PyTorch 1.8.0 with NVIDIA TITAN RTX GPU. All the student networks using distillation are trained using the ADAM optimizer with default hyperparameters in PyTorch. Unlike the previous work setting (200 \cite{fakd} or 300 \cite{lsfd} epochs), FACD loss is not sufficiently saturated. Therefore, the batch size and total epochs are set to 16 (same as in the previous works) and 600 epochs, respectively. The initial learning rate is set to \begin{math}2 \times 10^{-4}\end{math}, and is halved at 150 epochs. In addition, the patch size for training is set to \begin{math}48\times48\end{math} for network input, and default settings of data augmentation (e.g. horizontal flip, vertical flip, and random rotation) is applied. These experimental settings for reproduced results are equally applied to FAKD\cite{fakd} and LFD\cite{lsfd} which are main comparison works in this paper.

\begin{table}[h]
\centering
\caption{Network descriptions of teacher and student networks. T and S denote the teacher and student network respectively.}

\begin{tabular}{|c|cc|cc|}
\hline
\multirow{2}{*}{} & \multicolumn{2}{c|}{EDSR} & \multicolumn{2}{c|}{RCAN} \\ \cline{2-5} 
                  & T           & S           & T            & S          \\ \hline
Channel size      & 256         & 64          & 64           & 64         \\
ResBlocks         & 32          & 16          & 20           & 6          \\
ResGroups         & -           & -           & 10            & 10      \\
Params (M)        & 43          & 1.5         & 15.59        & 5.17       \\ \hline
\end{tabular}

\vskip 6pt

\label{table:configuration}
\end{table}

\begin{table*}[h]
\caption{Quantitative results (PSNR) measured by applying different FD methods on the student EDSR and RCAN network, as shown in \cref{table:configuration}. Note that the computational cost of inference stage for each FD method is the same. FAKD and FitNet indicate LSFD paper results, CSD* and LFD* indicate our reproduced results with our experimental settings. Except for them, the results in the table are taken from their respective paper. The best performance in the same experimental settings is marked in \textbf{bold}.}
\resizebox{\textwidth}{!}{%
\begin{tabular}{|c|c|c|c|c|c|c|c|c|c|c|c|}
\hline
\multicolumn{6}{|c|}{EDSR}                                                                                                                                                               & \multicolumn{6}{c|}{RCAN}                                                                                                                                                               \\ \hline
\multicolumn{1}{|c|}{Methods}     & \multicolumn{1}{c|}{Scale} & \multicolumn{1}{c|}{Set5}   & \multicolumn{1}{c|}{Set14}  & \multicolumn{1}{c|}{B100}   & \multicolumn{1}{c|}{Urban100} & \multicolumn{1}{c|}{Methods}     & \multicolumn{1}{c|}{Scale} & \multicolumn{1}{c|}{Set5}   & \multicolumn{1}{c|}{Set14}  & \multicolumn{1}{c|}{B100}   & \multicolumn{1}{c|}{Urban100} \\ \hline
\multicolumn{1}{|c|}{Teacher}     & \multicolumn{1}{c|}{x2}    & \multicolumn{1}{c|}{38.190} & \multicolumn{1}{c|}{33.857} & \multicolumn{1}{c|}{32.351} & \multicolumn{1}{c|}{32.873}   & \multicolumn{1}{c|}{Teacher}     & \multicolumn{1}{c|}{x2}    & \multicolumn{1}{c|}{38.271} & \multicolumn{1}{c|}{34.126} & \multicolumn{1}{c|}{32.390} & \multicolumn{1}{c|}{33.176}   \\ \hline
\multicolumn{1}{|c|}{Student}     & \multirow{7}{*}{x2}   & \multicolumn{1}{c|}{37.919} & \multicolumn{1}{c|}{33.439} & \multicolumn{1}{c|}{32.102} & \multicolumn{1}{c|}{31.728}   & \multicolumn{1}{c|}{Student}     & \multirow{7}{*}{x2}    & \multicolumn{1}{c|}{38.074} & \multicolumn{1}{c|}{33.623} & \multicolumn{1}{c|}{32.199} & \multicolumn{1}{c|}{32.317}   \\
\multicolumn{1}{|c|}{FAKD}        &     & \multicolumn{1}{c|}{37.976} & \multicolumn{1}{c|}{33.523} & \multicolumn{1}{c|}{32.156} & \multicolumn{1}{c|}{31.906}   & \multicolumn{1}{c|}{FitNet}      &    & \multicolumn{1}{c|}{38.132} & \multicolumn{1}{c|}{33.759} & \multicolumn{1}{c|}{32.253} & \multicolumn{1}{c|}{32.460}   \\

\multicolumn{1}{|c|}{LFD}         &     & \multicolumn{1}{c|}{37.984} & \multicolumn{1}{c|}{{33.547}} & \multicolumn{1}{c|}{32.156} & \multicolumn{1}{c|}{31.896}   & \multicolumn{1}{c|}{FAKD}        &    & \multicolumn{1}{c|}{38.164} & \multicolumn{1}{c|}{33.815} & \multicolumn{1}{c|}{32.274} & \multicolumn{1}{c|}{32.533}   \\

\multicolumn{1}{|c|}{LFD*}         &     & \multicolumn{1}{c|}{37.986} & \multicolumn{1}{c|}{{33.528}} & \multicolumn{1}{c|}{32.159} & \multicolumn{1}{c|}{31.935}   & \multicolumn{1}{c|}{LFD}         &    & \multicolumn{1}{c|}{38.178} & \multicolumn{1}{c|}{33.840} & \multicolumn{1}{c|}{32.296} & \multicolumn{1}{c|}{32.669}   \\

\multicolumn{1}{|c|}{LSFD}        &     & \multicolumn{1}{c|}{{37.991}} & \multicolumn{1}{c|}{{33.529}} & \multicolumn{1}{c|}{{32.157}} & \multicolumn{1}{c|}{{31.936}}   & \multicolumn{1}{c|}{LFD*}         &     & \multicolumn{1}{c|}{38.180} & \multicolumn{1}{c|}{33.851} & \multicolumn{1}{c|}{32.305} & \multicolumn{1}{c|}{32.681}   \\

\multicolumn{1}{|c|}{CSD*}       &     & \multicolumn{1}{c|}{38.001} & \multicolumn{1}{c|}{33.536} & \multicolumn{1}{c|}{32.160} & \multicolumn{1}{c|}{31.944}   & \multicolumn{1}{c|}{LSFD}        &   & \multicolumn{1}{c|}{{38.189}} & \multicolumn{1}{c|}{{33.882}} & \multicolumn{1}{c|}{{32.323}} & \multicolumn{1}{c|}{{32.704}}   \\

\multicolumn{1}{|c|}{FACD (ours)} &    & \multicolumn{1}{c|}{\textbf{38.043}} & \multicolumn{1}{c|}{\textbf{33.588}} & \multicolumn{1}{c|}{\textbf{32.188}} & \multicolumn{1}{c|}{\textbf{32.072}}   & \multicolumn{1}{c|}{FACD (ours)} &    & \multicolumn{1}{c|}{\textbf{38.242}} & \multicolumn{1}{c|}{\textbf{34.016}} & \multicolumn{1}{c|}{\textbf{32.334}} & \multicolumn{1}{c|}{\textbf{32.878}}   \\ \hline
\multicolumn{1}{|c|}{Teacher}     & \multicolumn{1}{c|}{x3}    & \multicolumn{1}{c|}{34.547} & \multicolumn{1}{c|}{30.435} & \multicolumn{1}{c|}{29.167} & \multicolumn{1}{c|}{28.470}   & \multicolumn{1}{c|}{Teacher}     & \multicolumn{1}{c|}{x3}    & \multicolumn{1}{c|}{34.758} & \multicolumn{1}{c|}{30.627} & \multicolumn{1}{c|}{29.309} & \multicolumn{1}{c|}{29.104}   \\ \hline
\multicolumn{1}{|c|}{Student}     & \multirow{7}{*}{x3}    & \multicolumn{1}{c|}{34.272} & \multicolumn{1}{c|}{30.266} & \multicolumn{1}{c|}{29.044} & \multicolumn{1}{c|}{27.959}   & \multicolumn{1}{c|}{Student}     & \multirow{7}{*}{x3}    & \multicolumn{1}{c|}{34.557} & \multicolumn{1}{c|}{30.408} & \multicolumn{1}{c|}{29.162} & \multicolumn{1}{c|}{28.482}   \\
\multicolumn{1}{|c|}{FAKD}        &     & \multicolumn{1}{c|}{34.356} & \multicolumn{1}{c|}{30.296} & \multicolumn{1}{c|}{29.066} & \multicolumn{1}{c|}{28.016}   & \multicolumn{1}{c|}{FitNet}      &    & \multicolumn{1}{c|}{34.570} & \multicolumn{1}{c|}{30.466} & \multicolumn{1}{c|}{29.184} & \multicolumn{1}{c|}{28.493}   \\
\multicolumn{1}{|c|}{LFD}         &    & \multicolumn{1}{c|}{34.348} & \multicolumn{1}{c|}{30.287} & \multicolumn{1}{c|}{29.068} & \multicolumn{1}{c|}{27.999}   & \multicolumn{1}{c|}{FAKD}        &    & \multicolumn{1}{c|}{34.653} & \multicolumn{1}{c|}{30.449} & \multicolumn{1}{c|}{29.208} & \multicolumn{1}{c|}{28.523}   \\
\multicolumn{1}{|c|}{LFD*}         &    & \multicolumn{1}{c|}{34.333} & \multicolumn{1}{c|}{30.301} & \multicolumn{1}{c|}{29.077} & \multicolumn{1}{c|}{28.022}   & \multicolumn{1}{c|}{LFD}         &     & \multicolumn{1}{c|}{34.657} & \multicolumn{1}{c|}{{30.525}} & \multicolumn{1}{c|}{29.224} & \multicolumn{1}{c|}{28.665}   \\
\multicolumn{1}{|c|}{LSFD}        &     & \multicolumn{1}{c|}{{34.384}} & \multicolumn{1}{c|}{{30.302}} & \multicolumn{1}{c|}{{29.077}} & \multicolumn{1}{c|}{{28.029}}   & \multicolumn{1}{c|}{LFD*}         &     & \multicolumn{1}{c|}{34.659} & \multicolumn{1}{c|}{{30.515}} & \multicolumn{1}{c|}{29.226} & \multicolumn{1}{c|}{28.672}   \\

\multicolumn{1}{|c|}{CSD*}       &    & \multicolumn{1}{c|}{34.378} & \multicolumn{1}{c|}{30.309} & \multicolumn{1}{c|}{29.082} & \multicolumn{1}{c|}{28.020} & \multicolumn{1}{c|}{LSFD}        &     & \multicolumn{1}{c|}{{34.666}} & \multicolumn{1}{c|}{30.510} & \multicolumn{1}{c|}{{29.226}} & \multicolumn{1}{c|}{{28.689}}   \\ 
\multicolumn{1}{|c|}{FACD (ours)} &    & \multicolumn{1}{c|}{\textbf{34.394}} & \multicolumn{1}{c|}{\textbf{30.333}} & \multicolumn{1}{c|}{\textbf{29.103}} & \multicolumn{1}{c|}{\textbf{28.125}}   & \multicolumn{1}{c|}{FACD (ours)} &    & \multicolumn{1}{c|}{\textbf{34.729}} & \multicolumn{1}{c|}{\textbf{30.563}} & \multicolumn{1}{c|}{\textbf{29.262}} & \multicolumn{1}{c|}{\textbf{28.818}}   \\ \hline
\multicolumn{1}{|c|}{Teacher}     & \multicolumn{1}{c|}{x4}    & \multicolumn{1}{c|}{32.385} & \multicolumn{1}{c|}{28.741} & \multicolumn{1}{c|}{27.661} & \multicolumn{1}{c|}{26.425}   & \multicolumn{1}{c|}{Teacher}     & \multicolumn{1}{c|}{x4}    & \multicolumn{1}{c|}{32.638} & \multicolumn{1}{c|}{28.851} & \multicolumn{1}{c|}{27.748} & \multicolumn{1}{c|}{26.748}   \\ \hline
\multicolumn{1}{|c|}{Student}     & \multirow{7}{*}{x4}   & \multicolumn{1}{c|}{32.102} & \multicolumn{1}{c|}{28.526} & \multicolumn{1}{c|}{27.538} & \multicolumn{1}{c|}{25.905}   & \multicolumn{1}{c|}{Student}     & \multirow{7}{*}{x4}    & \multicolumn{1}{c|}{32.321} & \multicolumn{1}{c|}{28.688} & \multicolumn{1}{c|}{27.634} & \multicolumn{1}{c|}{26.340}   \\
\multicolumn{1}{|c|}{FAKD}        &    & \multicolumn{1}{c|}{\textbf{32.138}} & \multicolumn{1}{c|}{28.547} & \multicolumn{1}{c|}{27.557} & \multicolumn{1}{c|}{25.972}   & \multicolumn{1}{c|}{FitNet}      &    & \multicolumn{1}{c|}{32.417} & \multicolumn{1}{c|}{28.716} & \multicolumn{1}{c|}{27.660} & \multicolumn{1}{c|}{26.406}   \\
\multicolumn{1}{|c|}{LFD}         &     & \multicolumn{1}{c|}{32.107} & \multicolumn{1}{c|}{28.524} & \multicolumn{1}{c|}{27.552} & \multicolumn{1}{c|}{25.962}   & \multicolumn{1}{c|}{FAKD}        &    & \multicolumn{1}{c|}{32.461} & \multicolumn{1}{c|}{28.750} & \multicolumn{1}{c|}{27.678} & \multicolumn{1}{c|}{26.422}   \\
\multicolumn{1}{|c|}{LFD*}         &     & \multicolumn{1}{c|}{32.099} & \multicolumn{1}{c|}{28.557} & \multicolumn{1}{c|}{27.546} & \multicolumn{1}{c|}{25.962}   & \multicolumn{1}{c|}{LFD}         &     & \multicolumn{1}{c|}{32.475} & \multicolumn{1}{c|}{{28.783}} & \multicolumn{1}{c|}{27.693} & \multicolumn{1}{c|}{{26.542}}   \\
\multicolumn{1}{|c|}{LSFD}        &    & \multicolumn{1}{c|}{32.107} & \multicolumn{1}{c|}{28.548} & \multicolumn{1}{c|}{{27.563}} & \multicolumn{1}{c|}{{25.980}}   & \multicolumn{1}{c|}{LFD*}         &     & \multicolumn{1}{c|}{32.479} & \multicolumn{1}{c|}{{28.774}} & \multicolumn{1}{c|}{27.688} & \multicolumn{1}{c|}{{26.547}}   \\

\multicolumn{1}{|c|}{CSD*}       &     & \multicolumn{1}{c|}{32.112} & \multicolumn{1}{c|}{{28.563}} & \multicolumn{1}{c|}{27.569} & \multicolumn{1}{c|}{25.998}   & \multicolumn{1}{c|}{LSFD}        &     & \multicolumn{1}{c|}{{32.497}} & \multicolumn{1}{c|}{28.711} & \multicolumn{1}{c|}{{27.699}} & \multicolumn{1}{c|}{26.525}   \\ 
\multicolumn{1}{|c|}{FACD (ours)} &    & \multicolumn{1}{c|}{{32.128}} & \multicolumn{1}{c|}{\textbf{28.580}} & \multicolumn{1}{c|}{\textbf{27.580}} & \multicolumn{1}{c|}{\textbf{26.029}}   & \multicolumn{1}{c|}{FACD (ours)} &     & \multicolumn{1}{c|}{\textbf{32.540}} & \multicolumn{1}{c|}{\textbf{28.810}} & \multicolumn{1}{c|}{\textbf{27.708}} & \multicolumn{1}{c|}{\textbf{26.606}}   \\ \hline
  
\end{tabular}%
}
\label{table:results}
\end{table*}

\subsection{Quantitative Results}
The quantitative result (PSNR) is shown in \cref{table:results}. FACD achieves the best performance on almost benchmark datasets and scale factors except for the Set 5 data set on EDSR x4. The average performance improvement of EDSR and RCAN is about 0.1dB over conventional FD. The performance improvement of RCAN is better than that of EDSR. The biggest difference between EDSR and RCAN is the presence or absence of feature-attention blocks. This means that it is effective to make the features more similar in RCAN using the feature attention scheme. In other words, the knowledge of teachers can be better utilized in RCAN than EDSR on feature domain.

\noindent \textbf{Impact on scale factor}: \cref{tab:improvement} summarizes the evaluation results of \cref{table:results}. PSNR improvement averages the difference between the FACD and the overall FD performance in the \cref{table:results}. As shown in \cref{tab:improvement}, we confirmed that the efficiency of performance improvement of FACD over other FD decreases as the scale factor increases. In general, the performance improvement efficiency of scale x2 is about two times better than that of scale x4. As the scale factor increases, the texture restoration becomes more difficult, and limits the upper bound of the performance of the teacher network. This means that the knowledge to be transmitted by the teacher is limited at a larger scale factor. 

\noindent \textbf{Performance on Urban100}: Benchmark datasets for SISR each have their own data characteristics. For instance, Set5 and 14 are samples of simple objects, and BSD100 has various characteristics from natural images to complex textures. Urban100 dataset has a variety and repetition of patterns and edges that arise from the complex architecture of buildings. As shown in \cref{table:results} and \cref{tab:improvement}, FACD has better quantitative performance, especially on the Urban100 dataset compared to other datasets. This means that our FACD has an advantage over other FD approach for texture restoration. Our FACD achieves 0.66dB, 0.34dB, and 0.17dB PSNR improvement over baseline student, FAKD, and LSFD on scale 2 in RCAN network, respectively.

\begin{table}[]
\caption{Evaluation results on average PSNR improvement over other FD approaches. Performance improvement over 0.1dB is marked in \underline{underlined}. }
\label{tab:improvement}
\resizebox{\columnwidth}{!}{%
\begin{tabular}{|c|c|c|c|c|c|}
\hline
Model                 & scale & Set 5    & Set 14      & B100  & Urban100    \\ \hline
\multirow{3}{*}{EDSR} & x2    & +0.059    & +0.056       & +0.031 & \underline {+0.149} \\
                      & x3    & +0.034    & +0.030       & +0.031 & \underline{+0.108} \\
                      & x4    & +0.016    & +0.033       & +0.025 & +0.056       \\ \hline
\multirow{3}{*}{RCAN} & x2    & +0.073    & \underline{+0.169} & +0.045 & \underline{+0.229} \\
                      & x3    & +0.089   & +0.072       & +0.050 & \underline{+0.211} \\
                      & x4    & +0.074    & +0.062       & +0.024 & \underline{+0.118} \\ \hline
\end{tabular}%
}
\end{table}

\noindent \textbf{Comparison with pruning methods}: To further demonstrate the effectiveness of our proposed FD methods, we additionally compared the evaluation results with SOTA pruning methods such as ASSL \cite{assl} and SRP \cite{srp}. FLOPs measure when the output image size is set to \begin{math}3\times1280\times720\end{math}. For a fair comparison, the configuration of the student model for KD was set to be as close as possible to the computational cost of each pruned EDSR sub-network model. As shown in \cref{tab:pruning}, FACD achieves 0.02 dB improvement in PSNR over the SOTA pruning methods at smaller network sizes.

\begin{table*}
\centering
\caption{PSNR (dB) results on the Set5 (x2) in the EDSR sub-network. Note that ASSL \cite{assl} and SRP\cite{srp} indicate the original paper results. The best performance is marked in \textbf{bold}.}

\label{tab:pruning}
\resizebox{\textwidth}{!}{%
\begin{tabular}{|cc|cc|cccc|c|}
\hline
\multicolumn{2}{|c|}{Params (K)}       & \multicolumn{2}{c|}{FLOPs (G)}       & \multicolumn{4}{c|}{Pruning}                                                                     & KD          \\ \hline
\multicolumn{1}{|c|}{Pruning} & Ours   & \multicolumn{1}{c|}{Pruning} & Ours  & \multicolumn{1}{c|}{Scratch} & \multicolumn{1}{c|}{L1-norm \cite{l1prune}}  &\multicolumn{1}{c|}{ASSL \cite{assl}} & SRP \cite{srp}  & FACD (ours) \\ \hline
\multicolumn{1}{|c|}{1101.8}  & 1087.2 & \multicolumn{1}{c|}{254.5}   & 251.1 & \multicolumn{1}{c|}{37.85}   & \multicolumn{1}{c|}{37.91}  &\multicolumn{1}{c|}{37.94} & 37.97 & \textbf{37.99}       \\ \hline
\multicolumn{1}{|c|}{681.1}   & 678.5  & \multicolumn{1}{c|}{157.5}   & 156.9 & \multicolumn{1}{c|}{37.81}   & \multicolumn{1}{c|}{37.81}   &\multicolumn{1}{c|}{37.91} & 37.89 & \textbf{37.93}       \\ \hline
\multicolumn{1}{|c|}{381.8}   & 378.8  & \multicolumn{1}{c|}{88.9}    & 88.2  & \multicolumn{1}{c|}{37.75}   & \multicolumn{1}{c|}{37.73}    & \multicolumn{1}{c|}{37.82} & 37.84 & \textbf{37.86}       \\ \hline
\multicolumn{1}{|c|}{154.2}   & 153.8  & \multicolumn{1}{c|}{36.5}    & 36.2  & \multicolumn{1}{c|}{37.56}   & \multicolumn{1}{c|}{37.58}   & \multicolumn{1}{c|}{37.70} & 37.71 & \textbf{37.73}       \\ \hline
\multicolumn{1}{|c|}{26.9}   & 26.2  & \multicolumn{1}{c|}{7.3}    & 7.1  & \multicolumn{1}{c|}{36.74}   & \multicolumn{1}{c|}{36.87}    & \multicolumn{1}{c|}{37.23} & 37.28 & \textbf{37.31}       \\ \hline

\end{tabular}}

\end{table*}

\subsection{Qualitative Results}
\begin{figure}
\centering
\includegraphics[width=8cm]{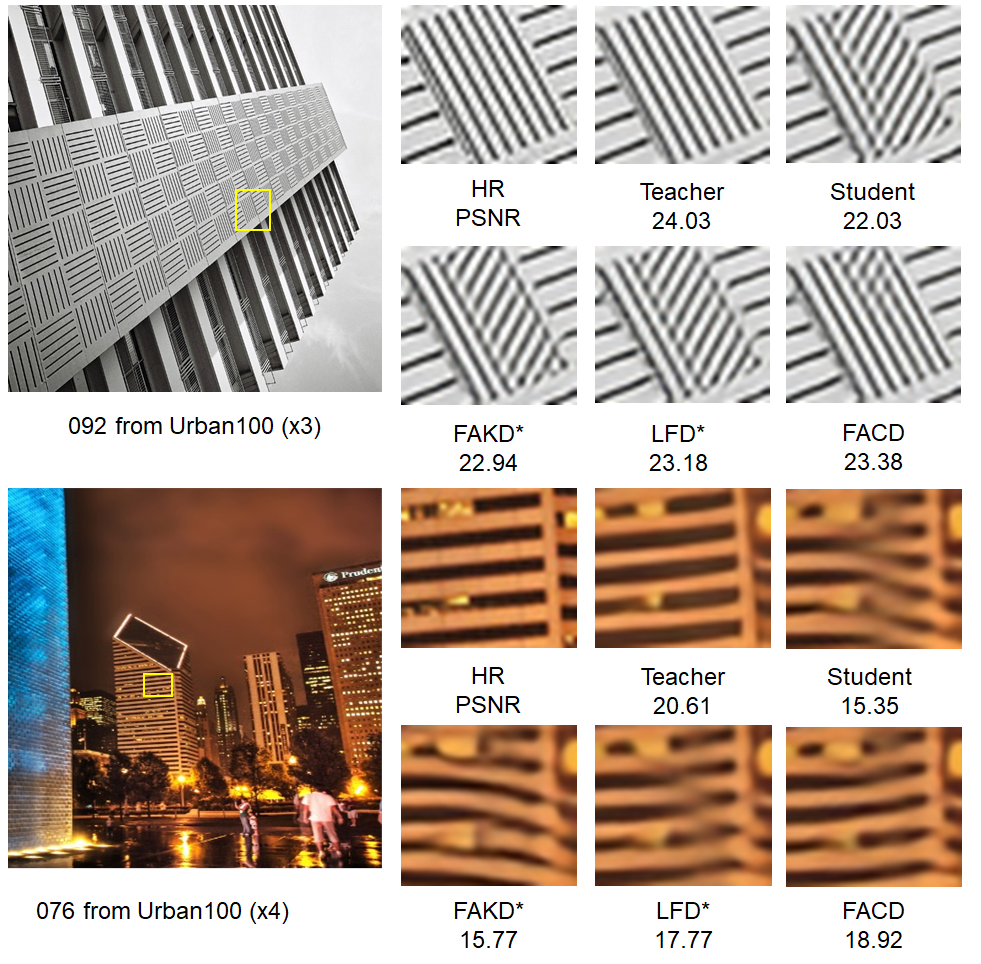}
\caption{Qualitative results on EDSR with scale x3 and x4. Noted that FAKD* and LFD* indicate our reproduced results with the same experimental settings. \label{figure_3}}
\end{figure}

As shown in \cref{figure_3}, we compare our FACD with other previous works on Urban100 benchmark dataset in terms of qualitative results. To compare the difference in the restoration of detailed patterns, we compared them to a relatively small cropped images. PSNR scores are calculated by considering only the cropped image. In general, as shown in the qualitative results, PSNR is proportional to the subjective image quality. We have clearly confirmed that our FACD achieves better PSNR and qualitative results than other FD approaches. Especially, for the restoration of textures (e.g. patterns), our FACD provides a clearer texture and is more similar to both the teacher and HR images than other FD methods. In the case of scale x3 images (092 from Urban 100), PSNR of FACD and LFD is not significantly different, but it shows a larger difference in the reconstruction of patterns that has been distorted by the conventional student networks. In other case (076 from Urban 100, x4), it shows a difference in the sharpness of the line and the straightness.

\section{Ablation Study} \label{abl}
This section demonstrates the effectiveness of the proposed FACD approach and we conduct an ablation study to analyze the contrastive loss, the effectiveness of each loss component, and adaptive distillation. 

\noindent \textbf{Impact on contrastive loss domains}: To show the effectiveness of contrastive loss in the feature domain, we compared the contrastive loss results in the feature and image domains. The composition of contrastive loss such as the formation of the equation is the same except for the applied domain. As shown in \cref{tab:racd}, our FCD achieves better results than ICD and CSD. Feature-domain contrastive distillation achieves 0.04dB PSNR improvement over the image-domain contrastive distillation. 

\begin{table}[]
\centering
\caption{Ablation study (PSNR) on the contrastive loss comparison between image and feature domains (x2, EDSR). FCD refers to a method that removes the adaptive scheme from the FACD. CSD* indicates our reproduced results with our experimental settings. }

\label{tab:racd}
\resizebox{\columnwidth}{!}{%
\begin{tabular}{|l|c|c|c|c|c|}
\hline
Methods &Distillation Domain   & \multicolumn{1}{l|}{Set5}      & \multicolumn{1}{l|}{Set14}     & \multicolumn{1}{l|}{B100}      & \multicolumn{1}{l|}{Urban100}  \\ \hline
ICD     &Image  & 37.983 & 33.558 & 32.159 & 32.020 \\ \hline
CSD* \cite{csd2021}     &VGG features of image   & 38.001 & 33.536 & 32.160 & 31.944 \\ \hline
FCD(ours) & Intermediate features & 38.025                         & 33.581                         & 32.183                         & 32.064  \\ \hline
\end{tabular}%
}
\end{table}

\noindent \textbf{Impact on contrastive loss formulation}: InfoNCE loss has been mainly used in contrastive learning on unsupervised learning \cite{infonce}. Similarity measure of contrastive loss in InfoNCE loss used the dot-product operation. In other hands, our FACD used the Euclidean distance loss as a similarity measure for contrastive loss. As shown in \cref{tab:infonce}, Contrastive loss with Euclidean loss achieves the 0.02dB PSNR improvement over the InfoNCE loss.

\begin{table}[h]
\centering
\caption{Ablation study (PSNR) of the different contrastive loss (\begin{math}L_{FACD}\end{math}) on EDSR x2.}

\label{tab:infonce}
\resizebox{\columnwidth}{!}{%
\begin{tabular}{|l|c|c|c|c|}
\hline
Methods    & \multicolumn{1}{l|}{Set5}      & \multicolumn{1}{l|}{Set14}     & \multicolumn{1}{l|}{B100}      & \multicolumn{1}{l|}{Urban100}  \\ \hline
FCD(InfoNCE)       & 38.005 & 33.556 & 32.173 & 32.028 \\ \hline
              
FCD(ours) & 38.025                         & 33.581                         & 32.183                         & 32.064  \\ \hline
\end{tabular}%
}
\end{table}

\noindent \textbf{Impact on each loss component}: In order to confirm the effect on the performance of each loss component, the loss component is turned on/off and tested. L1\_GT is the conventional Euclidean loss over GT images and LT\_T is the image domain distillation loss. The difference between FCD and FACD is the presence of adaptive distillation. The overall results are shown in \cref{tab:component}. Compared with the baseline model without distillation, our distillation approach achieved significant performance improvement on all benchmark datasets. 

\cref{tab:component} in the 4th row shows that training also works well for the independent application of distillation without L1\_GT. In particular, the FCD or FACD components show a larger performance improvement over the other loss components, as shown by the performance difference between the 1st row and the 3rd row. This means that the proposed FCD or FACD provide the knowledge from the teacher to the student network effectively. Finally, the combination of all loss components achieves the best evaluation results on various benchmark datasets.

\begin{table}[h]
\caption{Ablation study (PSNR) on the effectiveness of each loss component in the RCAN network (x4). The best performance is marked in \textbf{bold}.}
\label{tab:component}
\resizebox{\columnwidth}{!}{%
\begin{tabular}{|cccc|c|c|c|c|}
\hline
\multicolumn{4}{|c|}{Loss Component} & \multirow{2}{*}{Set5} & \multirow{2}{*}{Set14} & \multirow{2}{*}{B100} & \multirow{2}{*}{Urban100} \\ \cline{1-4}
L1\_GT    & L1\_T   & FCD   & FACD   &                       &                        &                       &                           \\ \hline
\checkmark         &         &       &        & 32.321                & 28.688                 & 27.634                & 26.340                    \\
\checkmark         & \checkmark       &       &        & 32.362                & 28.722                 & 27.657                & 26.402                    \\
\checkmark         &         & \checkmark     &        & 32.425                & 28.738                 & 27.669                & 26.507                    \\
          & \checkmark       & \checkmark     &        & 32.447                & 28.768                 & 27.671                & 26.524                    \\
\checkmark         & \checkmark       & \checkmark     &        & 32.492                & 28.774                & 27.689                & 26.562                    \\
\checkmark         & \checkmark       &     & \checkmark      & \textbf{32.540}                & \textbf{28.810}                 &\textbf{27.708}                & \textbf{26.606}                    \\ \hline
\end{tabular}%
}
\end{table}

\noindent \textbf{Impact on adaptive KD}: In this \cref{adaptive}, we describe the effectiveness of worse cases of the teacher networks. In order to confirm the effect of performance with worse samples of teacher networks, we compared the quantitative results of our FACD and FCD (FACD without an adaptive distillation approach). As shown in \cref{tab:fcd}, except for Set5 and 14 in scale 4 (EDSR), FACD achieves 0.02dB PSNR improvement over the FCD. This ablation study confirms the importance of the adaptive distillation approach. 

\begin{table}[h]
\centering
\caption{Ablation study (PSNR) on the effectiveness of adaptive distillation methods in the EDSR and RCAN networks. FCD indicates feature-domain contrastive distillation without adaptive distillation. The better performance is marked in \textbf{bold}.}

\label{tab:fcd}
\resizebox{\columnwidth}{!}{%
\begin{tabular}{|c|c|cccc|}
\hline
\multirow{3}{*}{Methods} & \multirow{3}{*}{Scale} & \multicolumn{4}{c|}{Model}                                                                           \\ \cline{3-6} 
                         &                        & \multicolumn{2}{c|}{EDSR}                                   & \multicolumn{2}{c|}{RCAN}              \\ \cline{3-6} 
                         &                        & \multicolumn{1}{c|}{B100}   & \multicolumn{1}{c|}{Urban100} & \multicolumn{1}{c|}{B100}   & Urban100 \\ \hline
FCD                      & \multirow{2}{*}{x2}    & \multicolumn{1}{c|}{32.183} & \multicolumn{1}{c|}{32.064}   & \multicolumn{1}{c|}{32.331} & 32.851   \\ \cline{1-1} \cline{3-6} 
FACD                     &                        & \multicolumn{1}{c|}{ \textbf{32.188}} & \multicolumn{1}{c|}{ \textbf{32.072}}   & \multicolumn{1}{c|}{ \textbf{32.334}} &  \textbf{32.878}   \\ \hline
FCD                      & \multirow{2}{*}{x3}    & \multicolumn{1}{c|}{29.096} & \multicolumn{1}{c|}{28.124}   & \multicolumn{1}{c|}{29.242} & 28.771   \\ \cline{1-1} \cline{3-6} 
FACD                     &                        & \multicolumn{1}{c|}{ \textbf{29.103}} & \multicolumn{1}{c|}{ \textbf{28.125}}   & \multicolumn{1}{c|}{ \textbf{29.262}} &  \textbf{28.818}   \\ \hline
FCD                      & \multirow{2}{*}{x4}    & \multicolumn{1}{c|}{27.578} & \multicolumn{1}{c|}{26.013}   & \multicolumn{1}{c|}{27.689} & 26.562   \\ \cline{1-1} \cline{3-6} 
FACD                     &                        & \multicolumn{1}{c|}{ \textbf{27.580}} & \multicolumn{1}{c|}{ \textbf{26.029}}   & \multicolumn{1}{c|}{ \textbf{27.708}} &  \textbf{26.606}   \\ \hline
\end{tabular}}
\end{table}

\noindent \textbf{Impact on feature attention}: In this paper, the three intermediate feature matching points are configured for fair FD comparison. In order to compare the effect of each feature, we compared the evaluation results of attention of each feature point. The only difference among three methods is the composition of the \begin{math}w_j\end{math} in \cref{eq:loss_FACD}. As shown in \cref{tab:attentionlayer}, FCD with our attention version (FAT) achieves the best PSNR score over the attention of other feature matching points. This means that due to the cascading architecture of the CNN-based SR network, the upper part has more influence on the distillation performance than the bottom part.

\begin{table}[h]
\centering
\caption{Ablation study (PSNR) of the feature attention on EDSR x2. FAA, FAB, and FAT indicate paying attention to the average, bottom, and top parts of features matching points, respectively. FAU is the version of this paper. The better performance is marked in \textbf{bold}.}

\label{tab:attentionlayer}
\resizebox{\columnwidth}{!}{%
\begin{tabular}{|l|c|c|c|c|c|c|}
\hline
Methods   & \multicolumn{1}{c|}{[\begin{math}w_1\end{math}, \begin{math}w_2\end{math},\begin{math}w_3\end{math}] } & \multicolumn{1}{c|}{Set5}      & \multicolumn{1}{l|}{Set14}     & \multicolumn{1}{l|}{B100}      & \multicolumn{1}{l|}{Urban100}  \\ \hline
FCD(FAA)   & [0.33, 0.33, 0.33]    & 38.007 & 33.579 & 32.177 & 32.058 \\ 
FCD(FAB)   & [0.20, 0.30, 0.50]   & 37.998 & 33.577 & 32.175 & 32.050 \\    
FCD(FAT)  & [0.50, 0.30, 0.20] & \textbf{38.025}        & \textbf{33.581}                   & \textbf{32.183}                 & \textbf{32.064}  \\ \hline
\end{tabular}%
}
\end{table}
\section{Conclusion}
In this paper, we propose the feature-domain adaptive contrastive distillation (FACD) scheme for efficient super-resolution models. In order to better transfer knowledge from the teacher networks, we proposed FD with the contrastive loss which is to maximize the mutual information of features between student and teacher networks. In addition, we found that inappropriate inference results of teacher networks decreased the efficiency of FD. To solve this issue, we propose the adaptive distillation scheme by rejecting the inaccurate teacher's knowledge transfer during the KD process. The student networks of EDSR and RCAN with FACD achieve state-of-the-art (SOTA) results in the KD for SISR. In particular, our FACD achieves excellent qualitative results in terms of texture reconstructions. We believe that the proposed FACD approach can be applied to other computer vision tasks (e.g. detection, segmentation, and denoising).


{\small
\bibliographystyle{ieee_fullname}
\bibliography{repq}
}

\end{document}